\documentclass[letterpaper, 10 pt, journal, twoside]{IEEEtran}

\IEEEoverridecommandlockouts                              

\usepackage{comment}

\usepackage[compact]{titlesec}

\titlespacing{\section}{0.5pt}{0.9ex}{1ex}
\titlespacing{\subsection}{0pt}{0.2ex}{0.2ex}
\titlespacing{\subsubsection}{0pt}{0.2ex}{0.2ex}

\usepackage{graphics} 
\usepackage[demo]{graphicx}
\usepackage{amsmath} 
\usepackage{amssymb}  
\usepackage{caption}
\usepackage{subcaption}
\usepackage{array}
\usepackage{tabularx}

\usepackage{xcolor}
\usepackage{cite}

\usepackage{lipsum}
\usepackage{afterpage}

\title{\LARGE \bf Implicit Learning of Scene Geometry from Poses for Global Localization }

\author{Mohammad Altillawi$^{1,2}$, Shile Li$^{1}$, Sai Manoj Prakhya$^{1}$, Ziyuan Liu$^{1}$, and Joan Serrat$^{2}$

\thanks{Accepted version. To appear in the IEEE Robotics and Automation Letters}
\thanks{© 2023 IEEE.  Personal use of this material is permitted.  Permission from IEEE must be obtained for all other uses, in any current or future media, including reprinting/republishing this material for advertising or promotional purposes, creating new collective works, for resale or redistribution to servers or lists, or reuse of any copyrighted component of this work in other works}

\thanks{$^{1}$M. Altillawi, S. Li, S. Prakhya, and Z. Liu are with the AI Robotics \& Simulation team, Huawei Munich Research Center, Germany.}

\thanks{$^{2}$M. Altillawi and J. Serrat are with the Computer Vision Center (CVC) and Department of Computer Science, Universitat Autònoma de Barcelona, 08193 Bellaterra (Cerdanyola del Vallès), Spain.}

\thanks{Digital Object Identifier (DOI): see top of this page.}
        }

\makeatletter
\newcommand{\setcaptype}[1]{\def\@captype{#1}}
\makeatother

\newsavebox{\tempbox}

\begin{document}

\maketitle

\savebox{\tempbox}{\begin{minipage}{\textwidth}
\setcaptype{figure}
\centering
\vspace{-0.7cm}
    \includegraphics[width=0.95\linewidth]{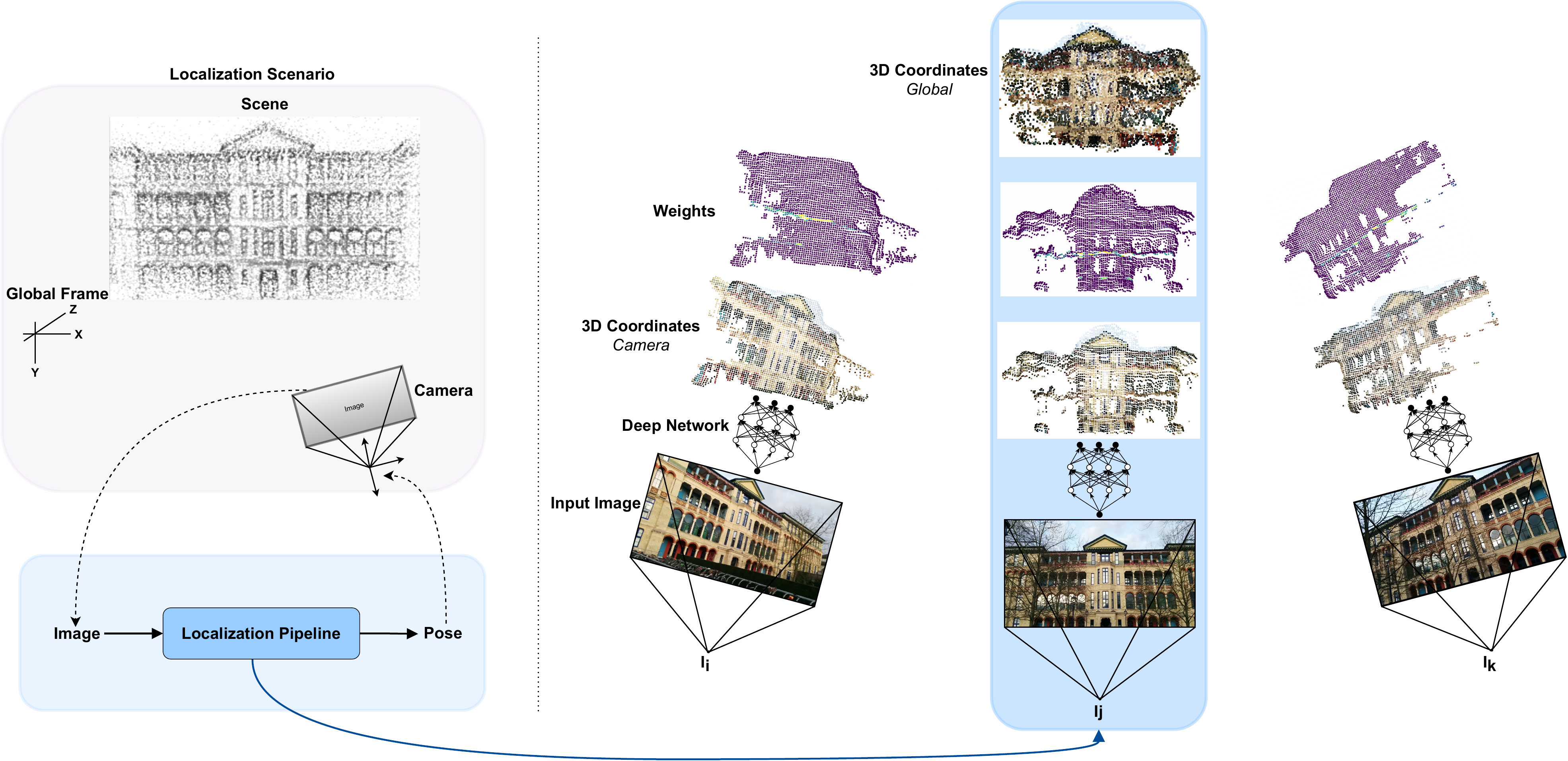}
    \caption{Illustration of our visual localization proposal on samples from Cambridge Landmarks (Hospital scene). Our method requires a set of images and the corresponding poses as the only labels for training. Left side: Given a single image, our method estimates the global pose of the camera in a given scene. Right side: we display the intermediate outputs of our proposal which are used to estimate the pose. For an input image, the proposed pipeline estimates two point clouds and a set of weights. The first point cloud represents the scene geometry (\textit{X, Y, Z coordinates}) in camera coordinate frame, while the second point cloud represents the scene geometry in global coordinate frame. These two point clouds, along with the predicted weights are used to estimate the camera's global pose. In the right side of Fig. 1, we visualize three sample input images, their corresponding indirectly estimated 3D scene representations (point clouds) and the weights. At the top, in the right side of Fig. 1, we can see only one 3D point cloud, which corresponds to three overlaid point clouds in the global coordinate frame, also estimated by our algorithm for the considered sample images. Though our method implicitly estimates 3D point clouds of the scene in local and global reference frames, it is not a mapping or 3D reconstruction algorithm, but a localization algorithm that implicitly learns and uses 3D scene geometry.
    }
    \label{teaser}
\end{minipage}}

\begin{figure}[t]
\rlap{\usebox\tempbox}
\end{figure}
\afterpage{\begin{figure}[t]
\rule{0pt}{\dimexpr \ht\tempbox+\dp\tempbox}
\end{figure}}


\begin{abstract}
Global visual localization estimates the absolute pose of a camera using a single image, in a previously mapped area. Obtaining the pose from a single image enables many robotics and augmented/virtual reality applications. Inspired by latest advances in deep learning, many existing approaches directly learn and regress 6 DoF pose from an input image. However, these methods do not fully utilize the underlying scene geometry for pose regression. The challenge in monocular relocalization is the minimal availability of supervised training data, which is just the corresponding 6 DoF poses of the images. In this paper, we propose to utilize these minimal available labels (.i.e, poses) to learn the underlying 3D geometry of the scene and use the geometry to estimate the 6 DoF camera pose. We present a learning method that uses these pose labels and rigid alignment to learn two 3D geometric representations (\textit{X, Y, Z coordinates}) of the scene, one in camera coordinate frame and the other in global coordinate frame. Given a single image, it estimates these two 3D scene representations, which are then aligned to estimate a pose that matches the pose label. This formulation allows for the active inclusion of additional learning constraints to minimize 3D alignment errors between the two 3D scene representations, and 2D re-projection errors between the 3D global scene representation and 2D image pixels, resulting in improved localization accuracy. During inference, our model estimates the 3D scene geometry in camera and global frames and aligns them rigidly to obtain pose in real-time. We evaluate our work on three common visual localization datasets, conduct ablation studies, and show that our method exceeds state-of-the-art regression methods' pose accuracy on all datasets.

\end{abstract}


\section{INTRODUCTION}
\IEEEPARstart{G}{lobal} camera re-localization has driven many computer vision applications in augmented/virtual reality and robotics. The problem definition, which is to obtain position and orientation (in metric units) of a camera from a single image in a previously mapped area, requires the availability of ground-truth poses for training.
 With training data that is composed of images and the corresponding poses, these methods follow a common learning process that maps the input image to 6 DoF directly in an end-to-end manner. This learning process goes as follows: the input image is passed to a network that encodes it into a latent vector. Through one or more regression layers, the latent vector is then mapped into a pose as two quantities, one for the translation and the other for the rotation. A pose loss is deployed to update the weights of the network during training. With a new input image at localization time, the network utilizes its memory formed by the learned weights to estimate the pose. Recent works \cite{posenet,posenet+,GposeNet,poselstm,svspose,hourglass,branchnet,mapnet,attloc, vipr, vidloc, vlocnet,PoGo_gnn, gnn, transformerPose, coordinet} build upon this scheme with different variations to constrain the weights to obtain a better pose estimate. This approach is attractive for several reasons. Firstly, it provides a pose directly in a single regression step.
 Secondly, the mapping from image to pose is fast; no classic feature matching is required, making it suitable for real-time applications. Thirdly, the whole pose estimation pipeline is saved in a compact form as network weights.

However, one limitation of this formulation is that it treats pose estimation as a regression problem and consequently, it strips out the geometry of the scene from pose estimation. This conditions the design of the deep network's last layer to be fully connected layers. This precludes incorporating geometric constraints such as depth or 3D scene coordinates. These can only be included in the loss terms in the training phase. Otherwise, they are explicitly required during inference (for example, 3D models from structure from motion).

In this work, we regard these limitations and propose a novel approach for global camera re-localization. Similar to other pose regression methods, our method is subject to the constraint of using minimalistic training data: a set of images with their intrinsics and the corresponding poses in a global reference frame. However, our pipeline does not estimate a pose by regression. Instead, it obtains geometric information of the scene which can be used, in turn, to estimate the pose geometrically; from pose labels only. The proposed pipeline learns 3D geometric representations (\textit{X, Y, Z} coordinates) of the scene as seen by the image, given guidance from ground-truth poses alone as shown in Fig. \ref{teaser}. It takes a single image and obtains 3D point cloud of the scene in two coordinate systems: camera frame and a global reference frame. To accomplish this, we utilize rigid alignment as a means to adjust the network weights in order to obtain two geometric maps. The rigid alignment module aligns the two clouds to obtain a pose. This pose is adjusted, through gradient descent while training, so as to match the ground-truth pose, thus, implicitly adjusting the two geometric representations (3D clouds) as well. 

In our end-to-end pipeline, deep learning is used to learn scene-specific geometric representations while we perform pose estimation in a geometric manner by parameter-free rigid alignment. During inference, it obtains a pose by rigidly aligning the two predicted 3D geometric representations. In contrast to pose regression, the proposed formulation allows the explicit use of additional constraints during training. We minimize re-projection errors in order to limit the deviation of the 3D map representation in the global frame from the corresponding 2D pixels. We complement that by constraining the deviation of the two 3D clouds according to ground-truth pose, which we refer to as consistency loss. This formulation results in higher localization accuracy than state-of-the-art regression methods. Additionally, we observe that our method can improve both position and orientation localization when finetuned with absolute position labels only. This is useful in applications where initially only a small set of training poses are available, but during operation, more data as partial geometric labels (position information from GPS) is available. 

In summary, our contributions are:
\begin{itemize}
\item We propose to utilize poses to train a network to learn geometric representations of a given scene implicitly. These are 3D coordinates in camera and a scene/global reference frame. For this goal, we utilize parameter-free and differentiable rigid alignment supervised by pose loss to guide the learning of scene geometry.
\item We also propose to employ additional loss terms, specifically, re-projection loss to constrain the learning of the 3D scene coordinates and introduce a consistency loss term that harmonizes the implicit geometric representations, according to the pose.
\item Apart from extensive evaluation on public datasets, we conduct ablation studies to evaluate the influence of the three different losses on our method's performance. The ablation studies show that our method can be finetuned using partial pose labels, i.e., with position information alone and still offers fairly good performance by improving both position and orientation accuracy.

\end{itemize}


\begin{figure*}[ht]
\centering
\includegraphics[scale=0.225]{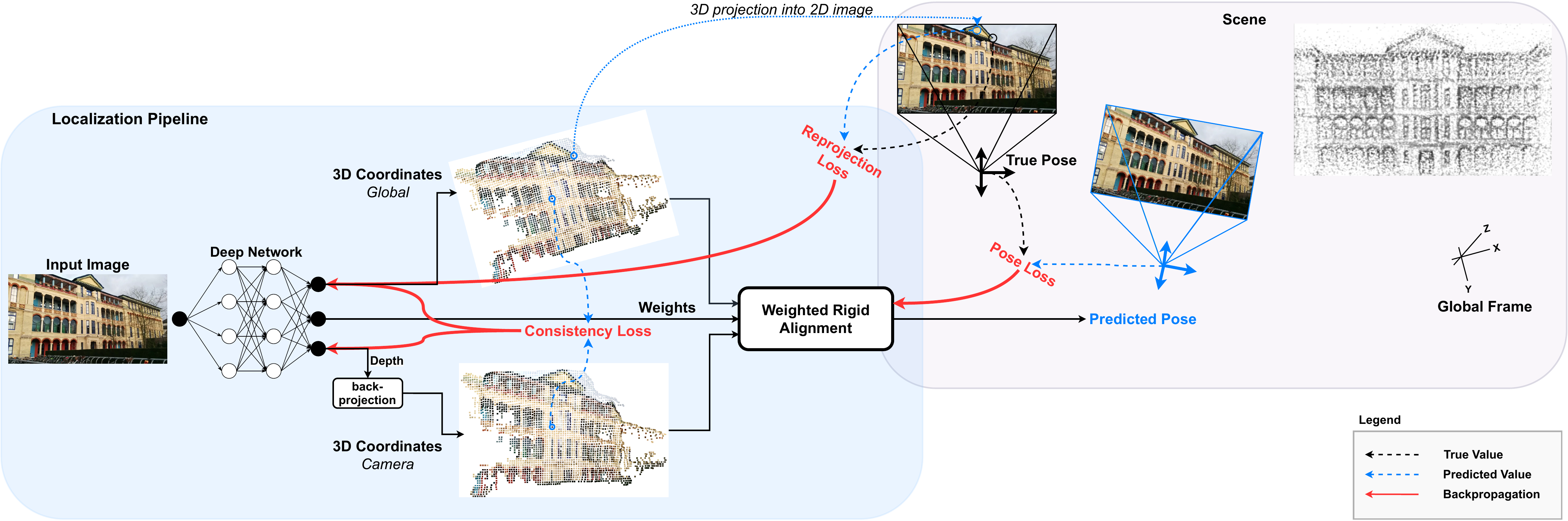} 
  \caption{A diagram of the proposed training method. Given images and the corresponding global pose labels, the network learns to indirectly (i.e., without explicit labels) estimate two 3D point clouds (one in global and the other in camera coordinate systems) and a set of weights. For the 3D coordinates in the camera coordinate system, the network predicts the depth, which is then back-projected to 3D space using Equation. \eqref{inv_proj}. A rigid alignment module aligns the two point clouds according to the weights to estimate pose. The loss terms employed for training are visualized in red.}
  \label{main_diagram}
\end{figure*}

\section{RELATED WORK}

The problem at hand is monocular global re-localization, which is to obtain metric poses (in meters and degrees) in a previously mapped area. Existing works utilize different sets of labels for training a localization pipeline. While some works utilize 3D geometric information such as 3D scene coordinates and depth, others utilize only pose labels. Similar to pose regression methods, our proposed method learns from pose labels only).

Initial works \cite{posenet,posenet+,GposeNet,poselstm,svspose,hourglass,branchnet,mapnet,attloc, vipr, vidloc, vlocnet,PoGo_gnn, gnn, transformerPose, coordinet} followed the regression approach and implemented different network architectures, learning strategies, and constraints on the learning process to reduce localization errors. In the following, we briefly review these methods and point out the coincidences and differences with respect to ours. PoseNetLSTM \cite{poselstm} utilizes LSTMs to reduce the dimensionality of the latent vector (that encodes the image) as a way to mitigate overfitting and obtain a more accurate pose estimate. PoseNetLearned \cite{posenet+} improves localization accuracy of their initial work PoseNet \cite{posenet} by addressing the issue of loss imbalance between the orientation and translation losses by learning weighting factors for these terms using homoscedatic uncertainty. PoseNetGeo \cite{posenet+} further improves the accuracy for some scenes by utilizing explicit 3D coordinate labels through re-projection loss in addition to pose loss. Though explicit 3D coordinate labels are used, the re-projection loss plays a limited role as it is used to constrain the image latent representation for regression instead of learning 3D scene geometry for pose estimation. Our proposal does not use 3D coordinate labels for training but still learns 3D geometry and uses it for pose estimation. Hourglass PN \cite{hourglass} replaces the initial architecture of PoseNet \cite{posenet} by a Resnet34 \cite{resnet} and complements it with up-convolutions to preserve the fine-grained information of the input image forming an Hourglass-like architecture. BranchNet \cite{branchnet} also adapts the architecture of PoseNet \cite{posenet} to account for the complex coupling between position and orientation. The split in network branches for position and orientation is performed at an earlier stage of the series of convolutions. Other works \cite{vidloc, mapnet, vlocnet, vipr} utilize sequences of images to gain additional source of training signal by imposing relative pose constraints that are obtained between consecutive camera frames.

AtLoc \cite{attloc} proposes a method based on attention to guide the network to output latent image representation that encodes robust objects and features for pose regression. It further utilizes a sequence of images to learn temporally consistent and informative features. CoordiNet \cite{coordinet} embeds pixel coordinates into the convolution operation to encode geometric feature locations into pose regression. It appends two additional channels that contain 2D pixel locations, to the input tensor before applying the convolution. MsTransformer \cite{transformerPose} proposes a transformer-based approach for localization. It utilizes image latent representations that are obtained from CNN, for processing by separate Transformers to regress position and orientation. Transformer encoders are used to aggregate activation maps with self-attention and decoders convert latent features and scene encoding into pose predictions. With the advances in graph neural networks (GNN), PoGo-Net \cite{PoGo_gnn} and GNNPose\cite{gnn} formulates the pose regression based on GNNs, naturally propagating information between different views for the benefit of pose regression.

Similar to previous regression works, we train from the available image-pose labels for the given datasets. In contrast, we propose using deep learning to implicitly learn representations of the 3D geometry of the scene instead of directly learning the pose regression. Accordingly, we solve for the pose by a single-step closed form solution through the rigid alignment of 3D scene representations. There are other works that use additional sensing modality such as LiDAR or labeled depth data or structure from motion results for training and/or inference  \cite{dsacstar,pixselect,blanton2022structure, adjacent_distant}. However, in this work, we solely focus on training with posed images only.


\section{METHOD}

\subsection{Overview}

Figure \ref{main_diagram} shows an overview of the proposed method. We propose to use the global camera pose $\mathbf{T}$ of a given input image $I$ as a label to guide the training of a deep neural network to obtain representations of the scene.

For that purpose, we define our localization pipeline to take a given image as input and yield two sets of 3D points, each in a different coordinate system. The first one is a set of 3D coordinates $G = { \{\hat{\mathbf{g}}_i, ..., \hat{\mathbf{g}}_M\}}$ in the global reference frame of the scene. These are predicted directly by the network. The second one is a set of 3D coordinates $C = { \{\hat{\mathbf{c}}_i, ..., \hat{\mathbf{c}}_M\}}$ in the camera frame. For the latter, the network predicts depth, which is then back-projected using intrinsics parameters to get 3D coordinates in camera frame. Inherently, the two 3D points clouds are matched via the image pixel coordinates.

Using rigid alignment, a pose $\hat{\mathbf{T}}$ can be estimated by aligning the two point clouds. We utilize Kabsch algorithm \cite{Kabsch} for this goal. It is differentiable, parameter-free, and obtains a closed-form solution in a single step. This makes the pipeline end-to-end trainable.

To account for imperfections in predictions, the network predicts a set of weights $W = { \{w_i, ..., w_M\}}$, which evaluates how much each 3D correspondence between point clouds from camera and global coordinate frame contributes to the rigid alignment. Given such correspondences, the weighted Kabsch algorithm \cite{Kabsch} is then applied to estimate the relative pose from camera coordinate system to the global coordinate system. Given $M$ 3D coordinates, this weighted minimization goal is defined as:
\vspace{-0.05cm}
\begin{equation}
\underset{\hat{\mathbf{R}},\hat{\mathbf{t}}} {\arg\min} \sum_{i}^{M}w_i||\hat{\mathbf{g}}_i - \hat{\mathbf{R}}\hat{\mathbf{c}}_i - \hat{\mathbf{t}}||_2,
\label{weighted}
\end{equation}
which can be described as follows: the translation $\hat{\mathbf{t}}$ of the pose is removed by centering both point clouds:

\[ \boldsymbol{\mu}_{\mathbf{g}} =  \frac{\sum_{i} w_i\hat{\mathbf{g}}_i}{\sum_{i} w_i},  \quad  \bar{\mathbf{G}} = \mathbf{G} - \boldsymbol{\mu}_{\mathbf{g}} \]

\[ \boldsymbol{\mu}_{\mathbf{c}} =  \frac{\sum_{i} w_i\hat{\mathbf{c}}_i}{\sum_{i} w_i},  \quad  \bar{\mathbf{C}} = \mathbf{C} - \boldsymbol{\mu}_{\mathbf{c}}. \]

Rotation $\hat{\mathbf{R}}$ and translation $\hat{\mathbf{t}}$ are then recovered with SVD as follows:
\vspace{-0.01cm}
\[\mathbf{U}\mathbf{S}\mathbf{V}^T = \mathrm{svd}(\bar{\mathbf{C}}^{T}W\bar{\mathbf{G}}) \]
\[s = \mathrm{det}(\mathbf{VU}^T) \]
\[\hat{\mathbf{R}} = \mathbf{V} \begin{pmatrix} 
	1 & 0 & 0 \\
	0 & 1 & 0\\
	0 & 0 & s \\
	\end{pmatrix} \mathbf{U}^T\]
\[\hat{\mathbf{t}} = -\hat{\mathbf{R}}\boldsymbol{\mu}_{\mathbf{c}} + \boldsymbol{\mu}_{\mathbf{g}}.\]

We apply a pose loss to guide the rigid alignment so that the network learns the 3D geometric representations. Given a ground-truth pose $\mathbf{T}$ with rotation $\mathbf{R}$ and translation $\mathbf{t}$ components, a cost function can be defined to minimize the difference between the estimated and ground-truth components. We define the loss as the summation of position loss and the rotation loss: 
\begin{equation}
    L_{pose} = L_{position} + L_{rotation},
    \label{pose_loss}
\end{equation}
where
\begin{equation}
    L_{position} =  \parallel  \mathbf{t} - \hat{\mathbf{t}} \parallel_2,
    \label{loc_loss}
\end{equation}
defines the position error between the computed translation $\hat{\mathbf{t}}$ and the actual translation $\mathbf{t}$ and 

\begin{equation}
    L_{rotation} = \cos^{-1}(\frac{1}{2}(\mathrm{trace}(\hat{\mathbf{R}}\mathbf{R}^{-1}) -1))
    \label{rot_loss}
\end{equation}
measures the angular error between computed rotation $\hat{\mathbf{R}}$ and the ground-truth rotation $\mathbf{R}$.

The predicted pose is adjusted by gradient descent that is guided, during training, by the pose loss Equation. \eqref{pose_loss}, to match the ground-truth pose. This process indirectly adjusts the two geometric representations by passing gradients through the differentiable rigid alignment. The proposed formulation allows for the inclusion of additional constraints that actively guide the optimization of the implicit 3D geometric representations from the poses. Consequently, we introduce a consistency loss to constrain the geometric predictions to be aligned according to the ground-truth pose. We first transform the 3D points $C$ from camera coordinate frame to global coordinate frame using the ground-truth pose. The consistency loss measures the error between the 3D points $G$ in global coordinate frame and the 3D points $C$ transformed from camera coordinate frame, using the ground-truth poses. We define the consistency loss as:
\vspace{-0.01cm}
\begin{equation}
L_{consistency} = \frac{1}{M} \sum_{i}^{M} \parallel  \hat{\mathbf{g}}_i - \mathbf{T}\hat{\mathbf{c}}_i \parallel_2,
\label{consistency}
\end{equation}

Rather than predicting the 3D coordinates directly, we can adjust the network to predict one quantity which is depth. Given depth, which forms the \textit{Z} coordinate in the camera perspective, the \textit{X} and \textit{Y} are obtained directly from the image pixels and depth, given camera intrinsics parameters. Accordingly, the 3D points $C$ in camera coordinate frame are obtained by back-projecting the depth according to:
\begin{equation}
\mathbf{\hat{c}}_i = \hat{d}_i \mathrm{\mathbf{K}^{-1}}\mathbf{u}_i,
\label{inv_proj}
\end{equation}
where $\mathbf{u}_i$, $\mathbf{K}$, $\hat{d}_i$, and $\mathbf{\hat{c}}_i$ denote the homogeneous pixel coordinates, the camera intrinsic matrix, the depth, and the corresponding point in the camera frame, respectively.

In addition, the 3D global coordinates are further constrained by utilizing a re-projection loss to minimize the error between the re-projection of the 3D global coordinates into the image frame and the 2D image pixels. It is defined as:
\vspace{-0.01cm}
\begin{equation}
L_{reprojection} = \frac{1}{M} \sum_{i}^{M} \parallel \mathbf{u}_i - \pi(\mathbf{T}\hat{\mathbf{g}}_i) \parallel_2,
\label{reproj}
\end{equation}
where $\pi$ projects points from the 3D global frame into the image frame.

With pose labels and the defined formulation, our method implicitly learns geometric representations of the scene. 
Given an image at inference, the proposed method estimates the scene's geometry and utilizes it for pose computation.

\begingroup
\setlength{\tabcolsep}{2.0pt}
\begin{table*}[t]
\caption{Comparison against state-of-the-art localization methods on Cambridge Landmarks \cite{posenet} and 7Scenes \cite{Shotton2013SceneCR} datasets.
First and second best results are marked in \textbf{bold} and \underline{underline} respectively. '\textunderscore' denotes unavailable results.}
\label{indoor_outdoor}
\begin{center}
\scriptsize
\begin{tabular}{l |c |c c c c c | c c c c c c c c}
& & &  & Cambridge & & & & & & 7Scenes & & & \\
Method & Backbone & College & Hospital & Shop & Church & Average & Chess & Fire & Heads & Office & Pumpkin & Kitchen & Stairs & Average\\
\hline

\hline


PoseNetGeo \cite{posenet+}  & GoogLeNet& 0.88/1.04&	3.20/3.29& 0.88/3.78& 1.57/3.32 & 1.63/2.86 & 0.13/4.48 & 0.27/11.3 &  0.17/13.0 &  0.19/5.55 &  0.26/4.75 &  0.23/5.35 &  0.35/12.4 & 0.23/8.12\\

PoseNetLSTM \cite{poselstm} &  GoogLeNet&	0.99/3.65 & 1.51/4.29 & 1.18/7.44 & 1.52/6.68 & 1.30/5.51 & 0.24/5.77 &  0.34/11.9 &  0.21/13.7 &  0.30/8.08 &  0.33/7.00 &  0.37/8.83 &  0.40/13.7 & 0.31/9.85 \\

GPoseNet \cite{GposeNet} &  GoogLeNet &	1.61/2.29 & 2.62/3.89 & 1.14/5.73 & 2.93/6.46 & 2.08/4.59 & 0.20/7.11 &  0.38/12.3 &  0.21/13.8 &  0.28/8.83 &  0.37/6.94 &  0.35/8.15 &  0.37/12.5 & 0.31/9.95\\

BranchNet \cite{branchnet} &  GoogLeNet & \textunderscore & \textunderscore & \textunderscore & \textunderscore & - & 0.18/5.17 &  0.34/8.99 &  0.20/14.2 &  0.30/7.05 &  0.27/5.10 &  0.33/7.40 &  0.38/10.3 & 0.29/8.32 \\

SVS-Pose \cite{svspose} &  VGGNet & 1.06/2.81 & 1.50/4.03 & 0.63/5.73 & 2.11/8.11 & 1.32/5.17 & \textunderscore & \textunderscore & \textunderscore & \textunderscore & \textunderscore & \textunderscore & \textunderscore & -\\

Hourglass PN \cite{hourglass} &  ResNet34 & \textunderscore & \textunderscore & \textunderscore & \textunderscore  & -& 0.15/6.17 &  0.27/10.8 &  0.19/11.6 &  0.21/8.48 &  0.25/7.01 &  0.27/10.2 &  0.29/12.5 & 0.23/9.54 \\

MapNet \cite{mapnet} &  ResNet34&	0.94/1.99 & 2.03/3.60 & 0.80/6.34 & 1.66/4.01 & 1.36/3.99 & 0.08/3.25 &  0.27/11.7 &  0.18/13.3 &  0.17/5.15 &  0.22/4.02 &  0.23/4.93 &  0.30/12.1 & 0.21/7.78\\

AtLoc \cite{attloc} &  ResNet34& \textunderscore & \textunderscore & \textunderscore & \textunderscore & -  & 0.10/4.07 &  0.25/11.4 &  0.16/11.8 &  0.17/5.34 &  0.21/4.37 &  0.23/5.42 &  0.26/10.5 & 0.20/7.56 \\

CoordiNet \cite{coordinet} &  ResNet34 & 0.80/1.22 & 1.43/2.86  & 0.73/4.69 & 1.32/4.10 & 1.07/3.22 & \textunderscore & \textunderscore & \textunderscore & \textunderscore & \textunderscore & \textunderscore & \textunderscore \\

CoordiNet \cite{coordinet} &  EffNet b3 & 0.70/0.92 & 0.97/2.08  & 0.69/3.74 & 1.32/3.56 & 0.92/2.58 & 0.14/6.7 & 0.27/11.6 &  0.13/13.6 &  0.21/8.6 &  0.25/7.2 &  0.26/7.5 &  0.28/12.9 & 0.22/9.72 \\

PoGO-Net \cite{PoGo_gnn} &  Not Applicable  &	-/0.94 & -/1.69 & -/2.40 & -/\textbf{2.12} & -/1.78 & -/1.72 & -/6.23 & -/7.34 & -/3.93 & -/3.56 & -/3.85 & -/7.88 & -/4.93\\

MsTransformer \cite{transformerPose} & EfficientNetB0 & 0.83/1.47 &  1.81/2.39  & 0.86/3.07 &  1.62/3.99 &  1.28/2.73 & 0.11/6.38 &  0.23/11.5 &  0.13/13.0  & 0.18/8.14 &  0.17/8.42 &  0.16/8.92 &  0.29/10.3 &  0.18/9.51 \\

GNNPose \cite{gnn} &  ResNet34 &	0.59/\textbf{0.65}& 1.88/2.78& 0.50/2.87& 1.90/3.29 & 1.22/2.40 & 0.08/2.82 &  0.26/8.94 &  0.17/11.41 &  0.18/5.08 &  0.15/2.77 &  0.25/4.48 &  0.23/8.78 &  0.19/6.32\\

\hline

Ours & ResNet34 & \underline{0.48}/0.84& \underline{0.67}/\underline{1.14}& \underline{0.47}/\underline{1.91}& \underline{0.90}/3.05& \underline{0.63}/\underline{1.74}  & \textbf{0.05}/\underline{1.86}& \textbf{0.14}/\textbf{4.15}& \textbf{0.09}/\textbf{5.29}& \underline{0.14}/\underline{3.61}& \textbf{0.11}/\underline{2.78}& \textbf{0.10}/\underline{2.66}& \underline{0.19}/\underline{4.14} & \textbf{0.12}/\underline{3.50}\\

 & MobileNetV3 & \textbf{0.46}/\underline{0.81} & \textbf{0.61}/\textbf{1.09}& \textbf{0.44}/\textbf{1.71}& \textbf{0.87}/\underline{2.88}& \textbf{0.60}/\textbf{1.62} & \textbf{0.05}/\textbf{1.42}& \underline{0.16}/\underline{4.57}& \textbf{0.09}/\underline{6.14}& \textbf{0.13}/\textbf{3.46}& \underline{0.12}/\textbf{2.48}& \textbf{0.10}/\textbf{2.51}& \textbf{0.17}/\textbf{3.48} & \textbf{0.12}/\textbf{3.44}\\

\hline
\end{tabular}
\end{center}
\end{table*}
\endgroup

The overall loss is then the weighted combination of the pose loss, the re-projection loss, and the consistency loss:
\begin{equation}
L_{total} = \lambda_p L_{pose} + \lambda_c L_{consistency} + \lambda_r l_{reprojection},
\label{total_loss}
\end{equation}
where $\lambda_p$, $\lambda_c$, and $\lambda_r$ are the losses weighting factors.



\section{RESULTS}
\subsection{Datasets}
We conduct our experiments on three common visual localization datasets. These are the outdoor Cambridge Landmarks \cite{posenet}, the indoor 7Scenes \cite{Shotton2013SceneCR}, and the indoor 12scenes \cite{12scenes} datasets. They exhibit different characteristics:

\textbf{Cambridge landmarks} is an outdoor re-localization dataset that covers different landmarks of several hundred or thousand square meters in Cambridge, UK. The provided reference poses are reconstructed from structure from motion.
The challenges in this dataset arise from variety of illumination changes due to weather and dynamic objects such as cars and pedestrians. In addition, the training set size is relatively small (few hundred of images).

\textbf{7Scenes} contains 7 scenes that depict difficult scenarios, such as motion blur, reflective surfaces, repeating structures, and texture-less surfaces. Several thousand frames with corresponding ground-truth poses are provided for each scene's train and test splits.

\textbf{12Scenes} consists of 12 sequences with challenges similar to 7Scenes dataset. However, compared to 7scenes, it covers larger indoor environments with smaller number of training images, about several hundred frames for each scene.

\subsection{Setup} \label{arch}
We resize the input images to a standard 480 px height and normalize them by mean and standard deviation. During training, we apply, on the fly, color jittering and random in-plane rotations in the range [-30°, 30°].

We adjust every backbone network that we use in our experiments to obtain three outputs. The first is a 3 channels output that corresponds to the \textit{X}, \textit{Y}, \textit{Z} coordinates in the global frame. The second is a one-channeled depth prediction. Depth is obtained from a Sigmoid function and is then scaled to a range of [0.1 10] for indoor scenes and [0.1 600] for outdoor scenes. These hyperparameters are generalizable and adjustable to specific scenes. The third is also of one channel, followed by a Sigmoid, which stores the weights that weigh the contribution of the correspondences to the rigid alignment.

For updating the network weights, we use Adam optimizer with $\beta_1 = 0.9$, $\beta_2 = 0.999$, $\epsilon = 10^{-8}$, and a weight decay of $5\times10^{-4}$. We train the whole architecture from scratch for 400 epochs with learning rate $10^{-4}$.
We set the weighting factors $\lambda_p$, $\lambda_c$, and $\lambda_r$ of Equation. \eqref{total_loss} to 1, 1, and 0.001 respectively to provide a balanced contribution to gradient updates. The low factor for the reprojection error aims to stabilize the training.

Following previous methods, we report localization errors as median translation (in meters) and median orientation (in degrees) errors for all the experiments below. For some experiments, we list as well the average localization errors that is computed on all scenes of the corresponding datasets.

\subsection{Results: comparison to previous methods}
In this section, we compare our proposed method against state-of-the-art. We consider the methods that utilize pose labels to train a deep network for global localization from a single image without including additional supervision signals or more sensory data. To marginalize improvements that may come out of using a recent backbone, we implement our method using ResNet34 backbone \cite{resnet}, which is adopted by previous methods. We report the median localization errors in Table \ref{indoor_outdoor}.
As listed, our method, with both backbones, obtains the lowest localization errors on all of the scenes, except for the rotation measurements on the college and church scenes. Even though our method does not rely on labeled 3D ground-truth coordinates, it surpasses PoseNetGeo \cite{posenet+}, which uses explicit 3D coordinates. PoseNetGeo \cite{posenet+} obtains pose by regression which doesn't directly incorporate these available geometric information. In contrast, our method uses poses to infer the geometry of the scene, which is directly embodied for pose estimation by rigid alignment. AtLoc \cite{attloc} implements attention to utilize informative regions of a given image for pose regression. On the contrary, our method obtains weighting factors that are directly used to minimize contributions from outliers, thus, improving localization accuracy. We show the benefit of these weights in section \ref{outlier_filtering_methods}. Previous methods also implement other strategies such as graph-neural-networks \cite{gnn, PoGo_gnn}, transformers \cite{transformerPose} relative poses supervision \cite{mapnet}, and LSTMs \cite{poselstm} to better encode the image representation for the task of pose estimation. However, their performances are leveled by regression incapability to utilize geometric quantities directly for localization. Our method uses pose labels to guide the network to learn certain geometric features, which, in return, are used to compute a pose. It lets the network, through pose targets, to freely choose the suitable geometric features that are best for localization. Besides using ResNet34 \cite{resnet} as a backbone, we implement our method using MobileNetV3 \cite{mobilenetv3} due to its efficiency. As shown in Tab. \ref{indoor_outdoor}, employing MobileNetV3 \cite{mobilenetv3} as the backbone, gives the best results across datasets. In the rest experiments of following sections, we adopt MobileNetV3 \cite{mobilenetv3} as our backbone while also providing a comparison with other backbones.

\begin{table}[ht!]
\caption{Ablation results of section \ref{ablations} on Cambridge Landmarks \cite{posenet}, 
 7Scenes \cite{Shotton2013SceneCR}, and 12Scenes \cite{12scenes} datasets.}
\label{Tab_ablation_coll7sc}
\setlength{\tabcolsep}{2.3pt}
\begin{center}
\scriptsize
\begin{tabular}{l | c c c | c c c}
\hline
& $L_{pose}$ & $L_{reproj}$ & $L_{consist}$ &  Cambridge & 7Scenes &  12Scenes \\
\hline
1 & \checkmark & & & 0.72/3.91 & 0.138/3.88 & 0.067/2.48\\

2 & \checkmark & \checkmark & &  0.71/2.89  &  0.119/3.49 & 0.063/2.41\\

3 & \checkmark & & \checkmark & 0.67/1.75 & 0.118/3.49 & 0.066/2.34\\

4 & \checkmark & \checkmark & \checkmark &  0.60/1.62 & 0.116/3.44 & 0.061/2.33\\
\hline
5 & & Input Resolution/4 & & 0.59/1.62 & 0.115/3.13 & 0.060/2.31\\

6 & & Input Resolution/8 & & 0.60/1.62 & 0.116/3.44 & 0.061/2.33\\

7 & & Input Resolution/16 & & 0.68/1.77& 0.134/4.15 & 0.068/2.65\\
\hline

8 & & Ours + 3D Coordinates & & 0.64/1.85& 0.131/4.03 & 0.064/2.57\\
9 & & Ours + Depth & & 0.60/1.62 & 0.116/3.44 & 0.061/2.33\\
\hline
 & & Backbone & Run-time &  &  & \\
\hline
10 & &  HRNetV2 \cite{hrnet} & 256 ms & 0.64/2.01& 0.131/3.78 & 0.068/2.61\\
11 & &  ResNet34 \cite{resnet} & 45 ms & 0.63/1.74 & 0.124/3.50 & 0.064/2.49\\
12 & & DenseNet121 \cite{densenet} & 90 ms & 0.62/1.69 & 0.121/3.52 & 0.062/2.43\\
13 & & MobileNetV3 \cite{mobilenetv3} & 14 ms & 0.60/1.62 & 0.116/3.44 & 0.061/2.33\\
\hline
\end{tabular}
\end{center}
\end{table}

\subsection{Results: ablation study} \label{ablations}
Here, we delve into our proposal by examining the effect of the employed three different losses, and the performance of our method with different output resolutions and backbones. We list the results in Tab. \ref{Tab_ablation_coll7sc}.

\textbf{Losses:} we evaluate the contribution of every loss in our proposed method. Our method utilizes pose labels to guide the network through a pose loss Equation. \eqref{pose_loss}. In addition, we apply consistency loss to align the two geometric representations (local and global) according to the input pose, and lastly a re-projection loss to align 3D global coordinates to 2D image pixels. The Tab. \ref{Tab_ablation_coll7sc} (rows 1 to 4) presents the averaged results over all sequences on considered datasets, with different combination of loss terms.

The results from Tab. \ref{Tab_ablation_coll7sc} (rows 1 to 4) show that adding the re-projection loss (row 2) or the consistency loss (row 3) to the pose loss (row 1) results in lower errors than when training the network with the pose loss only (row 1, rigid alignment module). The consistency loss is more effective than the re-projection loss in supporting the pose loss to reduce localization errors on outdoor scenes. On indoor scenes, they exhibit almost similar behavior.
Combining all the losses (row 4) obtains the lowest errors on all datasets.

\textbf{Resolutions:} we evaluate the performance of our proposed method with different output resolutions. In our experiments, the resolution of the output geometric representations is $1/8$ of the input resolution. Here, we change the output resolution by changing the stride parameter. We report the results of two additional down-sampling factors: 4 and 16 in Tab. \ref{Tab_ablation_coll7sc} (rows 5 to 7). For a down-sampling factor of 16, we observe a slight increase in localization errors compared to a down-sampling factor of 8. On average, down-sampling by a factor of 4 results in slight improvement in localization.

\textbf{Depth versus 3D Camera coordinates:} we can adjust the network to either obtain 3D coordinates in the camera frame directly or obtain the depth. For the latter, 3D camera coordinates can be computed by Equation. \eqref{inv_proj}. Rows 8 and 9 in Tab. \ref{Tab_ablation_coll7sc} suggest that learning just the depth results in a better localization performance. This constrains the 3D coordinates predictions and eases the learning process, so that the network learns one quantity rather than 3 quantities.

\textbf{Backbones:} while many backbones could be used to implement our method, we look into them from the perspective of localization accuracy, run-time and compactness. The rows 10 to 13 in Tab. \ref{Tab_ablation_coll7sc} show the results using different backbones together with the run-time. Being the most compact, that is, with the smallest number of parameters (3.7 million), MobileNetV3 \cite{mobilenetv3} obtains the best localization results with the fastest run-time (for a down-sampling factor of 8).

\begingroup
\setlength{\tabcolsep}{3.0pt}
\begin{table}[h]
\caption{Effect of different filtering methods (section \ref{outlier_filtering_methods}). Best results are marked in bold.}
\label{outlier}
\begin{center}
\scriptsize
\begin{tabular}{l |c c c c}
\hline
Method & College & Hospital & Shop & Church\\
\hline

Rigid + No Filtering & 0.55/0.87& 0.79/1.63& 0.59/2.32& 1.13/3.55 \\
Rigid + Filter Dynamic  & 0.53/0.86& 0.79/1.62& 0.56/2.27& 1.06/3.49\\
Rigid + Filter Dynamic + Others & 0.52/0.93& 0.76/1.51& 0.50/2.19& 1.01/3.46\\
PnP + RANSAC & 0.49/1.04& \textbf{0.58}/1.64& \textbf{0.41}/2.72& 1.45/4.72\\
Rigid + RNASAC & \textbf{0.46}/1.13& 0.60/1.73& 0.46/3.30& 1.12/3.58\\
Rigid + Weights [ours]& \textbf{0.46}/\textbf{0.81}& 0.61/\textbf{1.09}& 0.44/\textbf{1.71}& \textbf{0.87}/\textbf{2.88}\\

\hline
\end{tabular}
\end{center}
\end{table}
\endgroup
\subsection{Results: outliers filtering} \label{outlier_filtering_methods}
Our method estimates weighting factors for each 3D-3D correspondence. These account for imperfections in 3D coordinates predictions and aim to down-weigh 3D correspondences that lead to inferior localization results (i.e., outliers).
We conduct experiments to assess the impact of these weighting factors. Since our method relies on a single image for localization, we consider the following methods:

\textbf{Rigid + No Filtering:} we compute pose using rigid alignment without incorporating the obtained weights. That is, all predicted 3D correspondences are of equal importance.

\textbf{Rigid + Filter Dynamic:} we utilize semantics to filter out contributions from sources of outliers, mainly dynamic objects. Specifically, we utilize the recently released InternImage \cite{internimage} to segment the input image into semantic classes. We filter out 3D points that correspond to pixels of dynamic classes. These are pedestrians, cars, bicycles, and trucks.

\textbf{Rigid + Filter Dynamic + Others:} we complement the removal of dynamic points by pruning 3D points that correspond to semantic classes that are presumably inferior to localization such as sky and trees.

\textbf{Rigid + RANSAC:} we utilize a robust outlier filter by pairing the rigid alignment algorithm with a RANSAC scheme \cite{ransac}. We apply RANSAC with a maximum of 2000 iterations and an inlier threshold of 10 cm between corresponding 3D points. We use 10 correspondences for pose estimation.

\textbf{PnP + RANSAC:} our method uses poses to obtain 3D coordinates in the global coordinate system of the scene. Besides utilizing rigid-alignment, our method offers the flexibility to adopt perspective-n-point (PnP) algorithm for pose estimation using the 3D global coordinates and the corresponding 2D pixels. We compute pose using PnP \cite{p3p} from 4 correspondences using RANSAC \cite{ransac}. We set 2000 as maximum number of RANSAC iterations, with an inlier threshold of 10 pixels. 

\textbf{Rigid + Weights [ours]:} we compute pose using weighted rigid alignment where the weights are the ones obtained by our method. These are obtained directly with a forward pass of the network without further processing of the obtained 3D correspondences or any off-the-shelf outlier filter.

Tab. \ref{outlier} lists the results on Cambridge Landmarks \cite{posenet}. Pruning off 3D points that correspond to dynamic objects by semantic segmentation shows improvements in localization compared to utilizing all the 3D points. In addition to removing dynamic points, filtering out correspondences from non-informative regions like sky and trees obtains additional improvements. The hospital scene does not contain dynamic objects which justifies the same result that is obtained without filtering any 3D-3D correspondence. While semantics can be used to filter complete dynamic objects, it might not filter all sources of outliers. Besides, it is subject to segmentation errors and may filter many useful features. Pairing a robust outlier filter such as RANSAC \cite{ransac} with both PnP \cite{p3p} and rigid alignment shows further improvements in estimated translation. However, it leads to increased errors on the Church scene and elevated orientation errors overall. We trace this back to the reason that many 3D points may be inferior and that few of them are subject to inlier checks. The Church scene poses a difficult localization scenario where the camera moves 360° around the scene. To overcome this and improve further, it would require increasing the number of considered points, relaxing the inlier threshold and/or increasing the maximum number of iterations. In summary, utilizing RANSAC \cite{ransac} for filtering outliers requires adjusting hyperparameters for each scene besides imposing a run-time burden.

In contrast to the mentioned methods, using the set of weights that are obtained by our method delivers a consistent reduction in position and orientation localization errors. These weights form a prior about the useful 3D coordinates for localization. It considers all correspondences for localization, nevertheless, with different weights. We complement these quantitative observations by showing visualizations of the network outputs in Fig. \ref{our_predictions}. Inspecting the visualization, we observe that the network guesses the architecture of the scene from pose labels only. Furthermore, the heatmaps of the weights show that it focuses on a small subset of features mainly edges and corners, implying that the correct geometric predictions lay on these features. It obtains an estimation of the depth and 3D scene coordinates without supervised labels for these quantities.

\begingroup
\setlength{\tabcolsep}{2.0pt}
\begin{table*}[ht!]
\caption{Results of experiment of section \ref{additional_notes}. Median errors (meters/degrees) are reported. Improvements as a result of finetuning by $L_{position}$ over training on 1/3 of samples are marked by underlines.}
\label{Tab_thirddataset}
\begin{center}
\scriptsize
\begin{tabular}{c | c | c c c c | c c c c c c c c}
\hline
Method & Training Scheme & College & Hospital & Shop & Church & Chess & Fire & Heads & Office & Pumpkin & Kitchen & Stairs\\
\hline
&trained on all samples & 0.46/0.81 & 0.61/1.09& 0.44/1.71& 0.87/2.88& 0.05/1.42& 0.16/4.57& 0.09/6.14& 0.13/3.46& 0.12/2.48& 0.10/2.51& 0.17/3.48\\

Ours & trained on 1/3 of samples& 0.52/0.85 & 0.69/1.49 & 0.59/2.53 & 0.95/3.04 & 0.06/1.60 & 0.18/4.73 & 0.13/7.7 & 0.14/3.48 & 0.11/2.73 & 0.12/2.81 & 0.19/3.50\\

&finetuned with $L_{position}$ & \underline{0.50}/0.88 & \underline{0.68}/\underline{1.32} & 0.59/\underline{2.44} & \underline{0.93}/\underline{2.95} & \underline{0.05}/\underline{1.55} & 0.18/5.28 & 0.13/7.7 & \underline{0.12}/\underline{3.41} & 0.12/3.13 & \underline{0.11}/\underline{2.63} & \underline{0.18}/3.74\\

\hline

& trained on all samples&	0.94/1.99 & 2.03/3.60 & 0.80/6.34 & 1.66/4.01 & 0.08/3.25 &  0.27/11.7 &  0.18/13.3 &  0.17/5.15 &  0.22/4.02 &  0.23/4.93 &  0.30/12.1\\

MapNet \cite{mapnet}  & trained on 1/3 of samples&	1.12/6.10 & 3.14/7.81 & 1.29/8.76 & 2.65/6.54 &  0.12/4.75 &  0.30/11.51 &  0.18/14.02 &  0.20/6.20 &  0.21/5.33 &  0.27/5.76 &  0.37/11.53\\

& finetunned with $L_{position}$ & 1.25/93.02 & 3.03/139.79 & 1.34/95.01 & 2.63/53.50 &  0.12/38.98 &  0.31/24.68 &  0.18/20.28 &  0.19/41.52 &  0.22/35.62 &  0.26/32.00 &  0.37/24.23\\
\hline
\end{tabular}
\end{center}
\end{table*}
\endgroup

\begin{figure}[ht!]
\scriptsize
  \centering
  \setlength{\tabcolsep}{2.0pt}
  \begin{tabularx}{\textwidth}{ c  c  c  c}

    Input Image & Weights & Depth &  Scene Coordinates \\ 
    
    \includegraphics[width=0.111\textwidth]{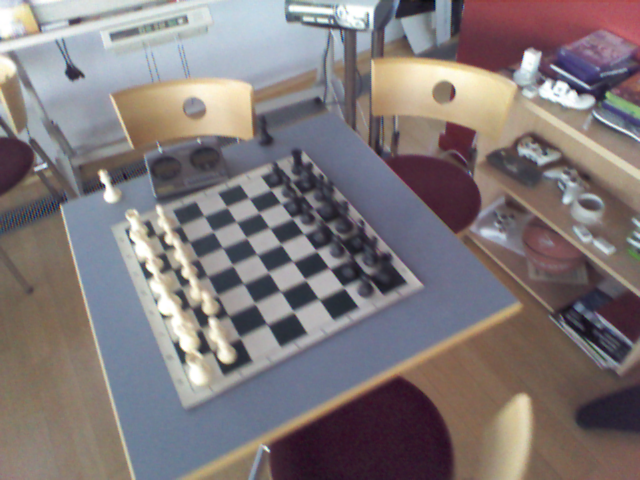}
    &    
    \includegraphics[width=0.111\textwidth]{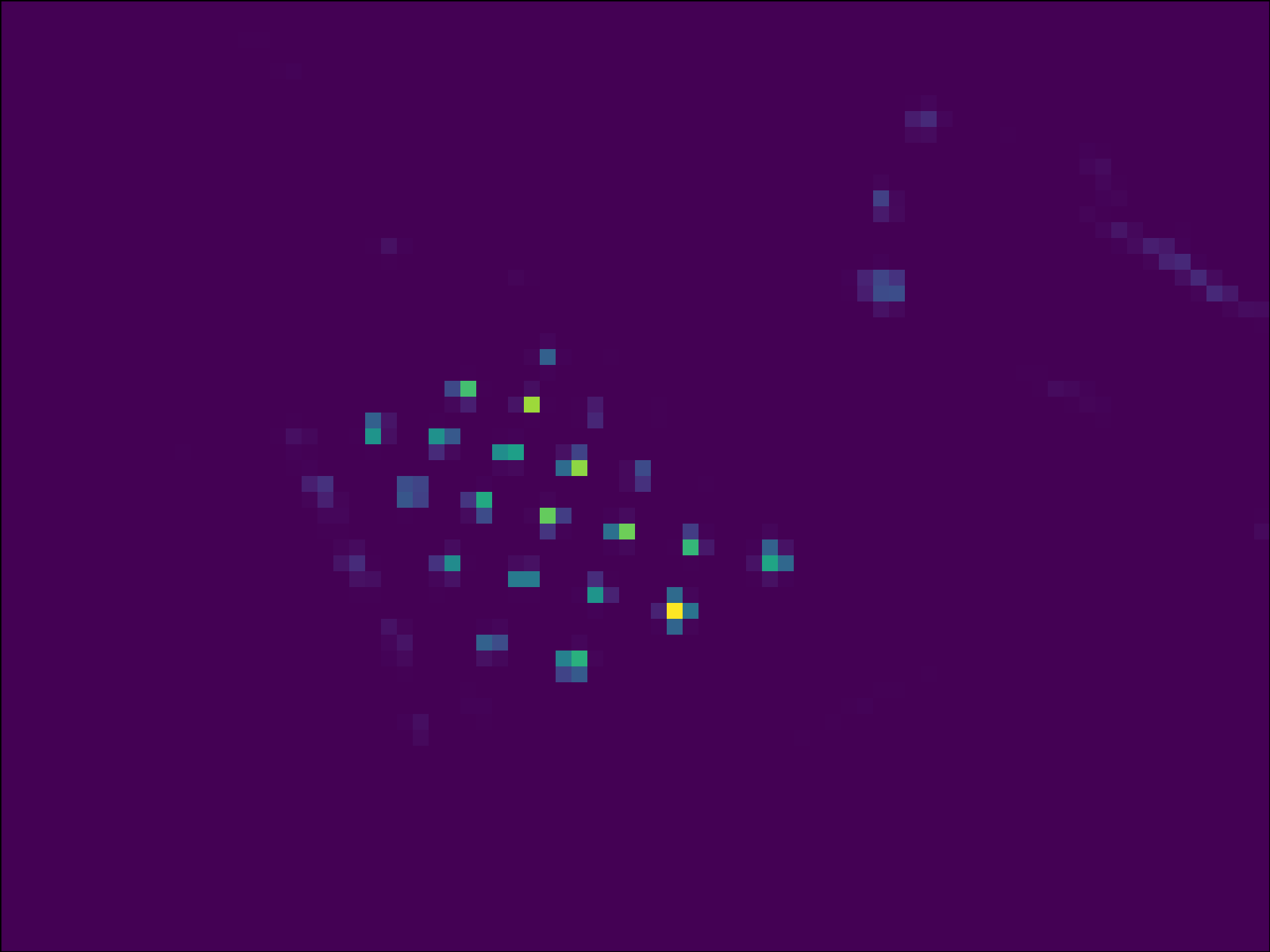}
    &
    \includegraphics[width=0.111\textwidth]{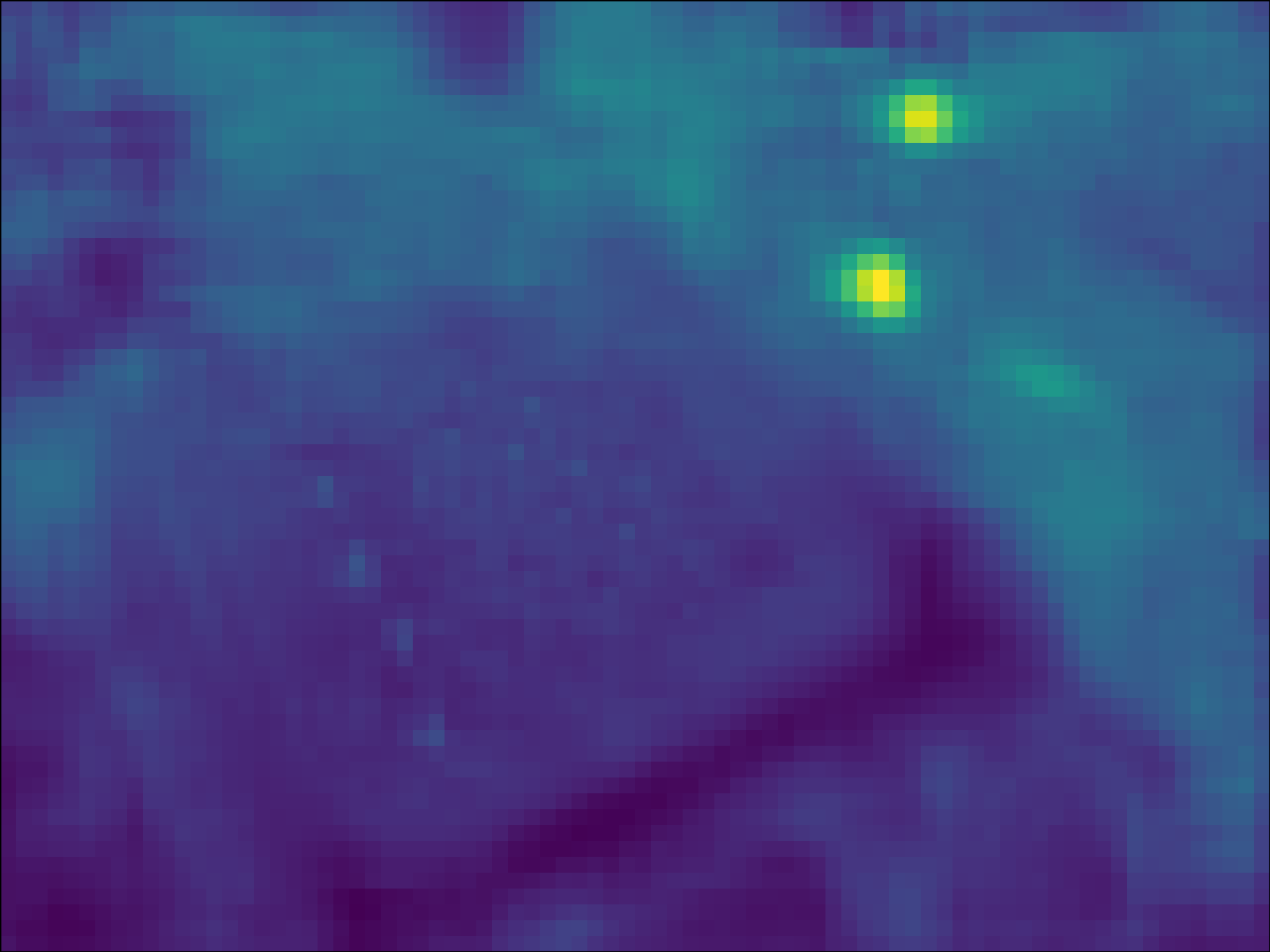}
    & 
    \includegraphics[width=0.111\textwidth]{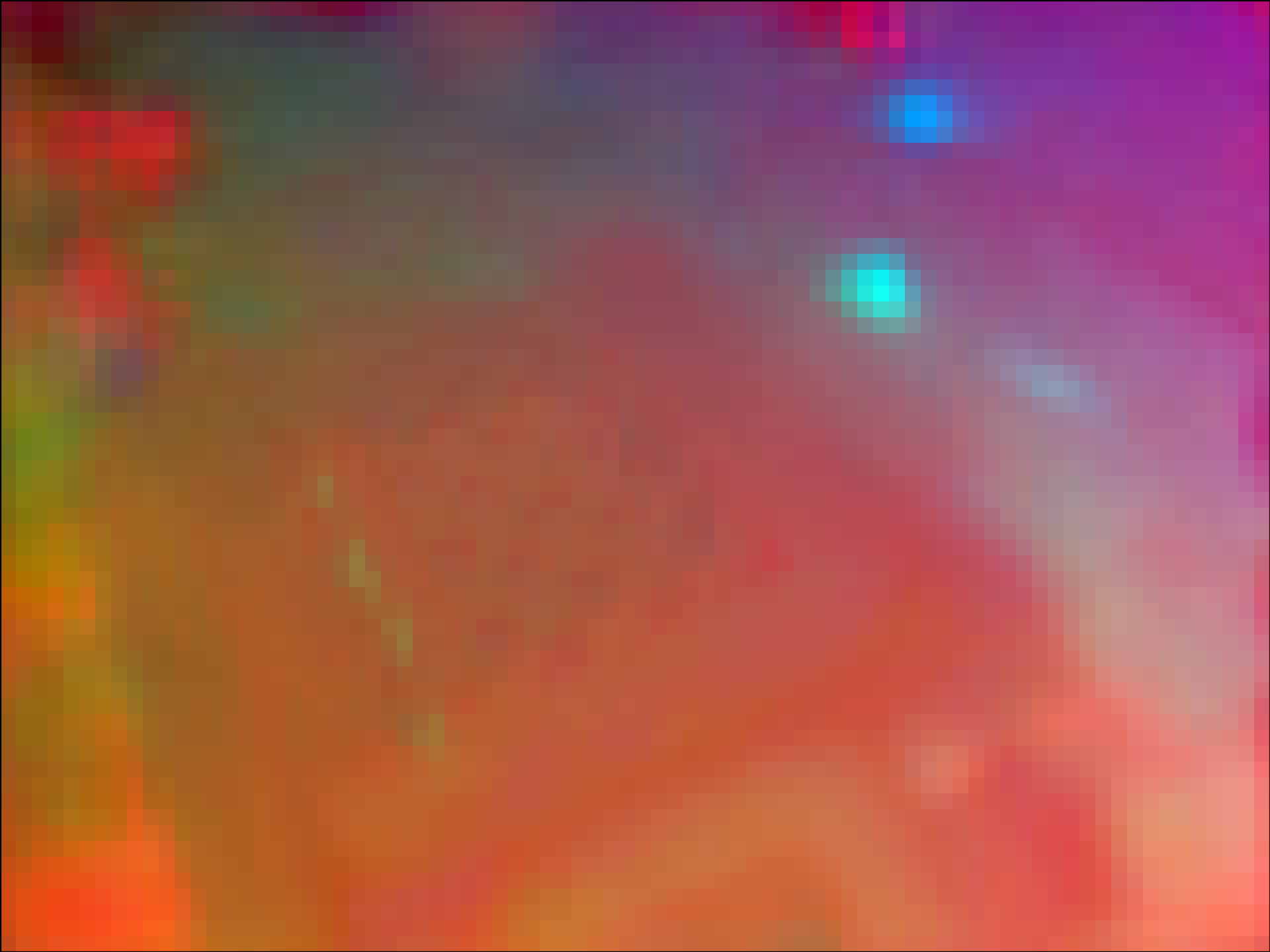}
    
    \\ 
    
    \includegraphics[width=0.111\textwidth]{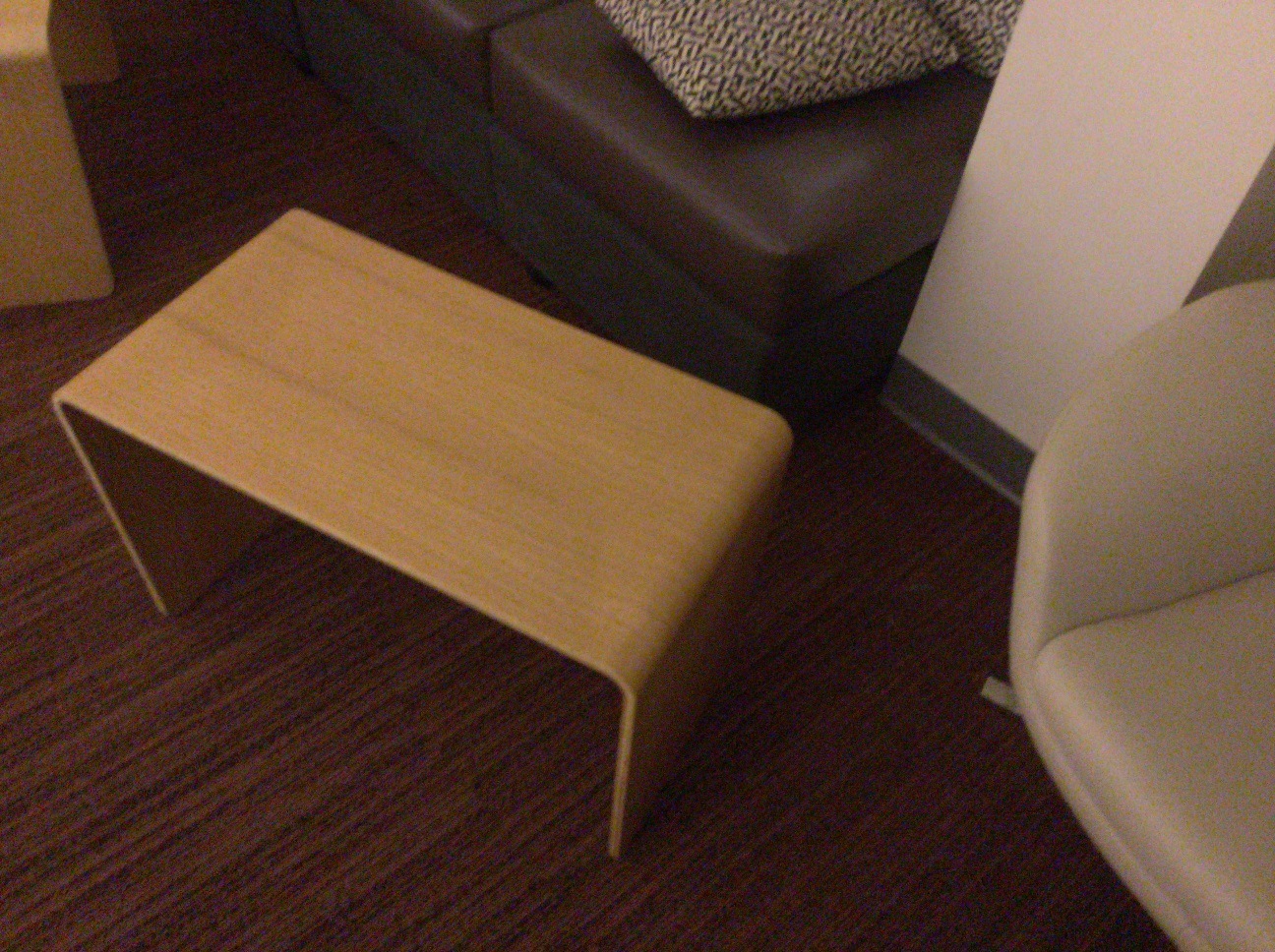}
    &    
    \includegraphics[width=0.111\textwidth]{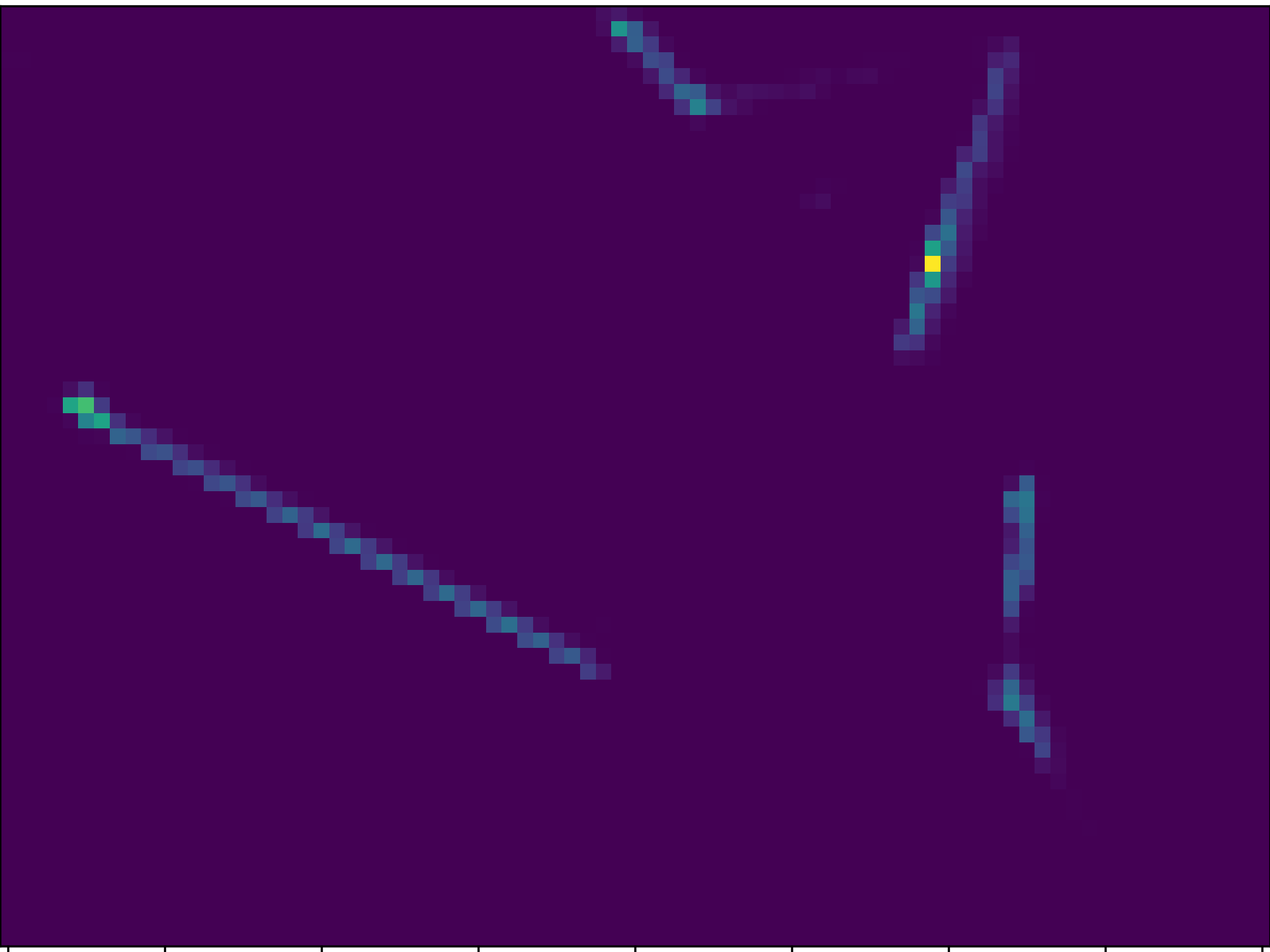}
    &
    \includegraphics[width=0.111\textwidth]{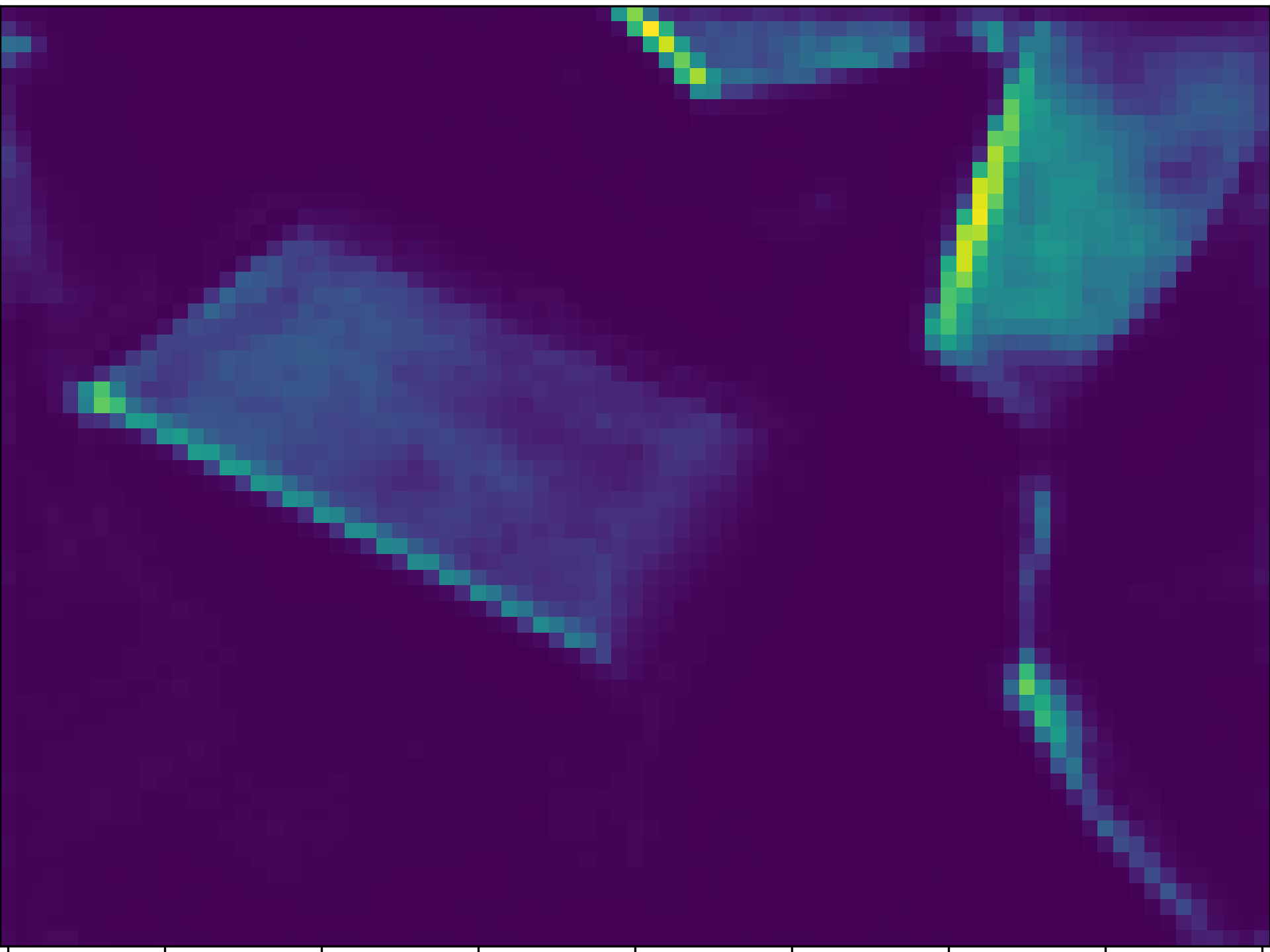}
    & 
    \includegraphics[width=0.111\textwidth]{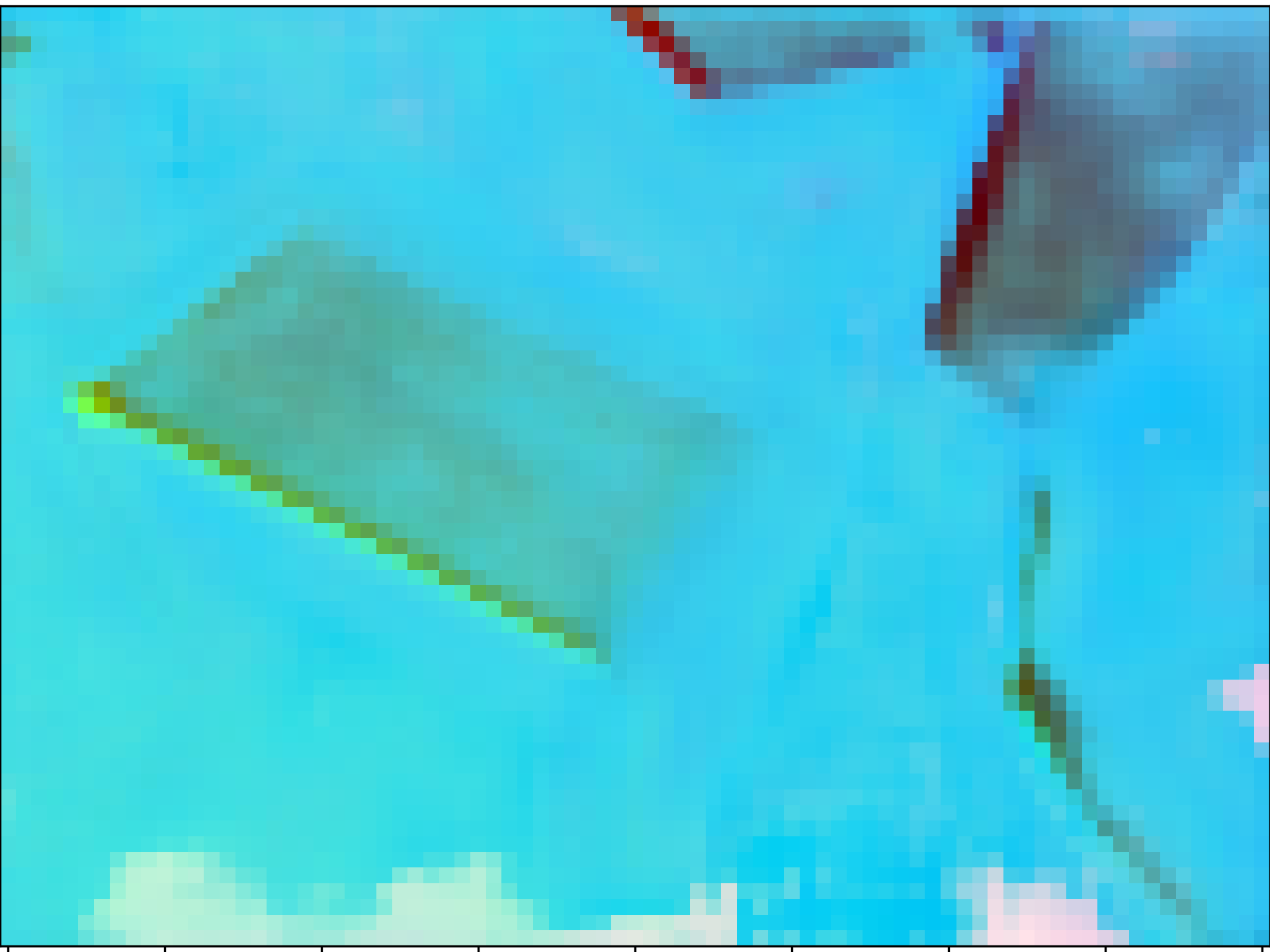}
    
    \\ 
    
  \end{tabularx}
  \caption{Visual samples of the predictions obtained by the network on samples from 7Scenes. For visualizing the 3D coordinates, we map the \textit{X}, \textit{Y}, \textit{Z} coordinates to RGB values.}\label{our_predictions}
\end{figure}

\subsection{Results: finetuning with position labels only} \label{additional_notes}

In this section, we test our method with limited training samples and also, explore the possibility of using only partial labels (only position information and not orientation). This partial position only labels can be easily available from GPS or can be available on the fly, using another sensor.
Testing with limited data is challenging for deep learning-based methods as they require more data for generalization. Accordingly, we sample a third of the dataset for training by taking every third training sample. The testing samples are kept in the original size as provided by the dataset. After training, we finetune the network for a few epochs (10-15) by feeding it with the other two-thirds of the training samples. However, we assume that these additional training samples have only partial geometric labels (translation only instead of 6 DoF poses). Thus, we apply only the translation loss for finetuning from Equation. \eqref{loc_loss}. The consistency (Equation. \eqref{consistency}) and re-projection (Equation. \eqref{reproj}) losses are excluded as they require full 6 DoF poses. In addition, we apply the same training scheme (training with absolute position labels) on one of state-of-the-art pose regression methods: MapNet \cite{mapnet}.
The results are listed in Table \ref{Tab_thirddataset}. As expected, the localization accuracy dropped when training with fewer samples. MapNet \cite{mapnet} yields a larger drop in accuracy than our method. Our method reports a little reduction in accuracy on indoor scenes. Finetuning the model given only position labels has reduced both the translation and rotation errors on some of the scenes. In contrast, MapNet \cite{mapnet} has significant increase in orientation errors when finetuned with position labels. The reason behind the better performance of our algorithm is driven by two correlated reasons. The first is the separation between image representations and pose estimation that is inherent to our algorithm, while the second being, the computation of the pose using non-learned and parameter-free rigid-alignment which updates both 3D point clouds from position supervision. The rigid alignment module passes gradients to all the network branches that predict the geometric quantities. Finetuning using only positional labels works well. While the accuracy did not reach that of training on the complete dataset, the improvements for both location and orientation given only translation labels open doors for further research in this direction.

\begin{table}[h]
\caption{Run-time analysis (section \ref{runtime}). Data Processing: time needed to run the network and obtain the 3D clouds. Pose Computation: time needed to obtain pose through rigid alignment. FPS: frames per second.}
\label{tab_runtime}
\begin{center}
\setlength{\tabcolsep}{3.0pt}
\scriptsize
\begin{tabular}{ c c c c c | c}
\hline
&Down-sampling & Output & Data & Pose& Total (ms)\\
&Factor & Resolution & Processing &  Computation & -FPS\\
\hline
&4 & $120\times160$  & 9.0 ms & 30.0 ms & 39.0 ms - 26 Hz \\
\rotatebox[origin=c]{90}{ours} & 8 & $60\times80$  & 9.2 ms & 4.8 ms & 14.0 ms - 71 Hz \\
&16 & $30\times40$  & 9.6 ms & 1.5 ms & 11.1 ms - 90 Hz \\

\hline
& & &  & Regression  & 12.5 ms - 80 Hz \\
\hline
\end{tabular}
\end{center}
\end{table}

\subsection{Results: run-time} \label{runtime}

In Table \ref{tab_runtime}, we report the run-time of our method for different down-sampling factors (resolutions) on an input with a standard resolution $480\times640$. We run our python implementation on a machine equipped with GTX Titan X and Intel Core i7-5960X CPU @ 3.00GHz. The results imply that our method localizes in real run-time. We also report the run-time of a minimal pose regression pipeline (PoseNet \cite{posenet}) using MobileNetV3 backbone \cite{mobilenetv3}. Some state-of-the-art regression methods further process the output of the network before pose regression by applying attention \cite{attloc}, graph neural networks \cite{gnn} and transformers \cite{transformerPose}, demanding additional run-time.

\section{CONCLUSION}
We presented a novel method for global 6 DoF pose estimation from a single RGB image.
The proposed work shares with most existing pose regression methods the same constraints, which are: train from a set of image-pose pairs, estimate a pose from a single image, save only the weights of the network, and obtain a pose in real run-time. However, our method obtains more accurate pose estimates as we have shown on common public datasets. The reason stems from the incorporation of scene geometry into pose estimation. The difficulty in achieving that, nonetheless, lies in the utilization of the only given labels (poses) to estimate this geometry and the use of the geometry for real run-time pose estimation. 
The main novelty of our method is the use of pose targets only to guide a deep neural network, through differentiable rigid alignment, to estimate the scene geometry without explicit ground-truth of this geometry at training time. The proposed method takes a single image and implicitly obtains geometric representations of the scene using only pose labels. These implicitly learned geometric representations are the 3D scene geometry (\textit{X}, \textit{Y}, \textit{Z} coordinates) in two reference frames: global and camera coordinate systems. We utilize a parameter-free and differentiable rigid alignment to pass gradients through a deep neural network to adjust its weights and continually learn these representations without explicit ground-truth labels. Besides pose loss, another novelty is that our method allows for the inclusion of additional learning losses as opposed to learning a localization pipeline by pose regression. We introduce a consistency loss to make the two geometric representations consistent with the geometric pose and a re-projection loss to constrain the 3D global coordinates to the 2D image pixels. Through extensive experiments, we show that the proposed method exceeds the localization accuracy of state-of-the-art regression methods and runs in real-time. As a final contribution, we show that the proposed formulation can utilize partial labels (instantaneous position labels only) to finetune a pre-trained model leading to improvements in both position and orientation localization. In future, we would like to leverage foundational models to generate embeddings and integrate them into our learned 3D representations to perform more accurate pose estimation using scene semantics.

\vspace{-0.17cm}
\bibliographystyle{IEEEtran}
\bibliography{root}
\end{document}